\documentclass{article}

\usepackage{microtype}
\usepackage{graphicx}
\usepackage{subcaption}
\usepackage{booktabs} %

\usepackage{hyperref}

\usepackage[preprint]{icml2026}

\usepackage{amsmath}
\usepackage{amssymb}
\usepackage{mathtools}
\usepackage{amsthm}
\usepackage{bm}

\renewcommand{\paragraph}[1]{{\vspace{1mm}\noindent \bf #1}.}

\usepackage[capitalize,noabbrev]{cleveref}

\theoremstyle{plain}

\theoremstyle{definition}

\theoremstyle{remark}

\usepackage[textsize=tiny]{todonotes}

\icmltitlerunning{Beyond Gaussian Bottlenecks: Topologically Aligned Encoding of Vision-Transformer Feature Spaces}

\begin{document}

\twocolumn[
  \icmltitle{Beyond Gaussian Bottlenecks: Topologically Aligned Encoding of Vision-Transformer Feature Spaces}

  \icmlsetsymbol{equal}{*}

  \begin{icmlauthorlist}
    \icmlauthor{Andrew Bond}{ku,kuisai}
    \icmlauthor{Ilkin Umut Melanlioglu}{kuisai,kuee}
    \icmlauthor{Erkut Erdem}{kuisai,hu}
    \icmlauthor{Aykut Erdem}{ku,kuisai}
\end{icmlauthorlist}

  \icmlaffiliation{ku}{Department of Computer Engineering, Koç University, Istanbul, Turkey}
  \icmlaffiliation{hu}{Department of Computer Engineering, Hacettepe University, Ankara, Turkey}
  \icmlaffiliation{kuisai}{KUIS AI Research Center, Istanbul, Turkey}
  \icmlaffiliation{kuee}{Department of Electrical and Electronics Engineering, Koç University, Istanbul, Turkey}

  \icmlcorrespondingauthor{Andrew Bond}{abond19@ku.edu.tr}

  \icmlkeywords{Machine Learning, ICML}

  \vskip 0.3in
]

\printAffiliationsAndNotice{}  %

\begin{abstract}
Modern visual world modeling systems increasingly rely on high-capacity architectures and large-scale data to produce plausible motion, yet they often fail to preserve underlying 3D geometry or physically consistent camera dynamics. A key limitation lies not only in model capacity, but in the latent representations used to encode geometric structure. We propose S$^2$VAE, a geometry-first latent learning framework that focuses on compressing and representing the latent 3D state of a scene, including camera motion, depth, and point-level structure, rather than modeling appearance alone. Building on representations from a Visual Geometry Grounded Transformer (VGGT), we introduce a novel type of variational autoencoder using a product of Power Spherical latent distributions, explicitly enforcing hyperspherical structure in the bottleneck to preserve directional and geometric semantics under strong compression. Across depth estimation, camera pose recovery, and point cloud reconstruction, we show that geometry-aligned hyperspherical latents consistently outperform conventional Gaussian bottlenecks, particularly in high-compression regimes. Our results highlight latent geometry as a first-class design choice for physically grounded visual and world models.

\end{abstract}

\section{Introduction}

Modern computer vision has increasingly shifted from pixel-level representations to high-dimensional feature spaces produced by large-scale Vision Transformers (ViTs)~\cite{vit}. Recent architectures further enrich these representations by embedding explicit geometric information, such as camera motion, depth, and 3D structure, directly into the Transformer’s hidden states. While these features are highly expressive, their dimensionality makes them impractical, necessitating aggressive compression into a compact latent space. 

A central but often implicit assumption in this process is that standard latent variable models, particularly Gaussian VAEs, are well matched to the geometry of transformer features. In practice, this assumption breaks down. Due to the extensive use of Layer Normalization and dot-product attention, ViT features concentrate on approximately hyperspherical manifolds, where semantic identity is encoded primarily through angular relationships rather than Euclidean magnitude. Compressing such features with a Gaussian VAE introduces a fundamental geometric mismatch: the model is forced to map data that lies on a thin spherical shell into a dense Euclidean volume. This mismatch manifests in several failure modes, including posterior collapse under strong compression, radial noise that perturbs angular semantics, and gradual semantic drift in reconstructed features. These effects are mild at low compression ratios but become pronounced precisely in the regimes that matter most, i.e. high compression and downstream geometry-sensitive tasks such as depth estimation, camera pose regression, and point cloud reconstruction.

Despite these issues, Gaussian latents remain the default choice, largely due to their mathematical convenience rather than empirical suitability. This raises a key question: \emph{Is the observed degradation under compression an unavoidable consequence of information loss, or is it amplified by a latent geometry that is fundamentally misaligned with Transformer features?}

To address this question, we introduce \textbf{S$^\mathbf{2}$VAE}, a geometry-aligned variational autoencoder that uses a product of hyperspherical latent distributions to respect the directional geometry of ViT feature representations under compression. Instead of a multivariate Gaussian bottleneck, we employ a \textit{product of Power Spherical distributions}, enforcing hyperspherical structure directly in the latent space. By decomposing the latent into multiple lower-dimensional spheres, the model maintains numerical stability while introducing a compositional inductive bias that facilitates the separation of distinct geometric attributes. Crucially, this alignment removes the need for the model to implicitly learn spherical constraints, allowing compression to focus on preserving semantically meaningful directions.

We show that this geometry-aligned bottleneck substantially reduces the failure modes observed with Gaussian VAEs. The resulting latent representation supports strong reconstruction fidelity even at high compression ratios and integrates naturally with task-specific geometric losses. As a result, downstream performance is preserved not merely because the model is expressive, but because the latent geometry is matched to the structure of the features it compresses.

Our contributions are threefold:
\begin{enumerate}
    \vspace{-1.2mm}
    \item We provide an empirical diagnosis of the mismatch between Gaussian latent assumptions and the intrinsic geometry of ViT features, and show that this mismatch leads to measurable degradation under compression.
    \vspace{-1mm}
    \item We propose a geometry-aligned latent bottleneck based on a product of hyperspherical distributions, enabling stable and differentiable compression of transformer features while preserving directional semantics.
    \vspace{-1mm}
    \item We demonstrate that latent alignment matters: compared to Gaussian VAEs, our approach consistently yields superior reconstruction quality and downstream geometric performance, particularly in high-compression regimes.
\end{enumerate}

\section{Related Work}
\subsection{Latent Variable Models for Non-Euclidean Data}
While standard VAEs assume Euclidean space and data, there have been much work into similar models for other types of data. The foundational work on this field is \cite{hyperspherical_vae}, which introduced the original Hyperspherical VAE formulation, based on the von Mises-Fisher (vMF) distribution. Meanwhile, \cite{power_spherical} introduced the power spherical distribution for VAEs, which helped resolve the instability issues and inefficient sampling of the vMF distribution. Additionally, \cite{explore_vae} explored other types of manifold-valued latent variables, focusing on the case of VAEs for Lie groups. While focused on simple experiments, they show that choosing the correct manifold for the data can strongly affect the latent representation. Finally, there is also some work which has explored the phenomenon of posterior collapse in VAEs, a common problem in training these models \cite{posterior_collapse}.
\subsection{Geometry of Transformer Features}
Much of the analysis of the geometry of transformer features has come from analysis of LLMs. \cite{pretrained_sentence_embeddings} showed that certain pretrained LLMs always induce non-smooth anisotropic semantic spaces, emphasizing the importance of focusing on certain directions. Meanwhile, \cite{contextual_word_representations} showed something similar, that in certain layers of LLMs, the anisotropy can be so bad that the cosine similarity between embeddings of two random words can be almost perfect. \cite{spherical_text_embedding} also argued that directional similarity is more important, and proposed a method to learn text embeddings in spherical space. 

\subsection{Visual Geometry and World Models}
Prior work has explored reconstructing visual signals from Vision Transformer representations, primarily with the goal of recovering pixel-level appearance or enabling generative modeling \cite{rae}. In contrast, our work focuses on representations that are explicitly grounded in 3D geometry rather than appearance. We build on geometry-centric foundation models such as VGGT \cite{vggt}, which are trained to encode camera motion, depth, and point-level structure directly in their hidden states. 

Recent extensions of this line of work have begun to address dynamic scenes, for example by disentangling pose and geometry across time \cite{page4d}. While these approaches demonstrate the expressiveness of geometry-aware transformers, they do not address how such high-dimensional geometric representations can be efficiently compressed or structured in a latent space. Our work complements these efforts by focusing on geometry-aligned latent representations that preserve directional and structural information under compression.

\section{Preliminaries}
\subsection{Visual Geometry Grounded Transformer (VGGT)}
VGGT is a geometry-first foundation model. Unlike standard ViTs which are trained on semantic tasks, VGGT is explicitly trained to understand the 3D structure of the world, by explicitly requiring the model to reconstruct a wide range of geometric primitives, such as camera poses, depth maps, and point clouds. Through this training, the model learns to produce tokenized representations that are grounded in physical space. These features are high-dimensional ($2048$-dimensional per layer) and typically are extracted across multiple depths of the model to capture both local geometric detail and scene context.

For our purposes, we utilize the intermediate features of a pretrained VGGT model as the primary input signal. These features are highly anisotropic, and due to the use of Layer Normalization, lie primarily on hyperspherical shells. While these tokens contain the necessary information to reconstruct complex 3D scenes, their density and high dimensionality make them tricky for use in downstream generative tasks. Consequently, the VGGT features are an ideal candidate for task-aware compression that can distill these multi-layer features into a unified latent state.

\subsection{Variational Autoencoders (VAEs)}
VAEs~\cite{vae} are a class of generative models that learn a compressed latent representation $\textbf{z}$ of an input signal $\textbf{x}$ by approximating the posterior distribution $p(\textbf{z}|\textbf{x})$. The architecture consists of an encoder $q_{\phi}(\textbf{z}|x)$ which maps the input to the parameters of a latent distribution, and a decoder $p_{\theta}(\textbf{x} | \textbf{z})$ which reconstructs the input from a sampled latent vector. The model is trained by maximizing the Evidence Lower Bound (ELBO), which consists of a reconstruction term and a regularization term based on the KL Divergence between the approximate posterior and a prior $p(\textbf{z})$. 
\begin{equation}
\mathcal{L}_\text{ELBO} = \mathbb{E}_{q_\phi(\mathbf{z}|\mathbf{x})}[\log p_\theta(\mathbf{x}|\mathbf{z})] - D_\text{KL}(q_\phi(\mathbf{z}|\mathbf{x}) | p(\mathbf{z}))
\end{equation}
In standard VAE implementations, the latent space is assumed to be Euclidean, and the prior distribution is defined as a multivariate Gaussian $\mathcal{N}(0, I)$. This choice allows for a simple reparametrization trick during training: $\textbf{z} = \mu + \sigma \odot \epsilon$ with $\epsilon \sim \mathcal{N}(0, 1)$. However, the variational framework is not limited to Euclidean geometry. The choice of the latent manifold and the corresponding prior $p(\textbf{z})$ are critical when determining the model's inductive bias. As we demonstrate below, when the input data resides on a non-Euclidean manifold such as a sphere, the standard Gaussian prior can introduce mismatches in training, which can be resolved by adopting hyperspherical latent distributions.

\section{Methodology}
\begin{figure*}
    \centering
    \includegraphics[width=0.9\linewidth]{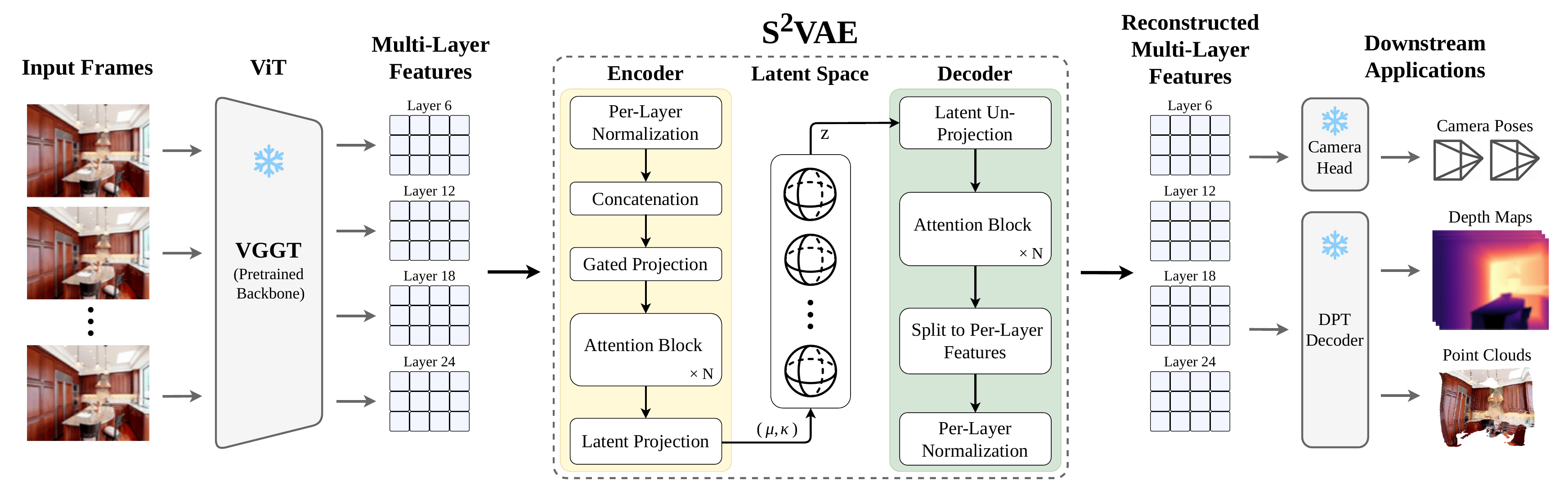}
    \caption{\textbf{Overview of the proposed S$^\mathbf{2}$VAE framework.} Input frames are first processed by a frozen ViT backbone (VGGT in this example) to extract multi-layer feature representations. Features from each layer are normalized independently, concatenated, and mapped to a shared hidden space via a gated projection. The encoder then applies a stack of self-attention blocks and predicts the parameters $(\mu,\kappa)$ of a hyperspherical latent distribution, modeled as a product of spheres, from which latent variables $\textbf{z}$ are sampled using spherical reparameterization. These latents are subsequently un-projected, refined through additional self-attention blocks, and projected back to the original feature dimensionality before being split into reconstructed layer features. The reconstructed features can be consumed by the original decoder heads to recover geometry-centric outputs such as camera pose, depth, and 3D point clouds.}
    \label{fig:overview}
\end{figure*}

Figure~\ref{fig:overview} presents an overview of the proposed architecture. 
Input frames are processed by a ViT backbone (VGGT in this illustration) to extract multi-layer features, which are individually normalized, concatenated, and projected into the VAE hidden space. 
After several self-attention blocks, the encoder predicts the $\mu$ and $\kappa$ parameters of a hyperspherical latent distribution; sampled latents are then decoded to reconstruct features that can be consumed by downstream task-specific heads.

\subsection{Hyperspherical Variational Autoencoder}
To effectively compress high-dimensional ViT features, we propose a Hyperspherical Variational Autoencoder architecture utilizing a hyperspherical latent space. While standard VAEs employ a multivariate Gaussian prior $\mathcal{P} = \mathcal{N}(0, I)$, we empirically show that this is sub-optimal for features produced by models like VGGT and DINO. These features typically reside on high-dimensional directional manifolds where magnitudes of features are approximately constant within a layer, due to the training methods and normalizations. A Gaussian prior, which couples direction and magnitude, introduces unnecessary radial noise. In contrast, hyperspherical distributions allow the model to focus explicitly on the angular semantics of the feature space.

\subsection{Dimensionality Reduction and Adaptation}
Before reaching the latent bottleneck, the input features $\textbf{x} \in \mathbb{R}^{D_{in}}$ are projected into a lower-dimensional hidden space. In the case where we choose to combine features from multiple layers together, we concatenate all the layers together along the feature dimension, and project down into the hidden dimension. We use a gated projection, given by
\begin{equation}
\mathbf{h} = \sigma(\mathbf{W}_g \mathbf{x} + \mathbf{b}_g) \odot (\mathbf{W}_h \mathbf{x} + \mathbf{b}_h)
\end{equation}
where $\textbf{h} \in \mathbb{R}^H$ represents the hidden state, and $H$ is typically chosen to be 512 or 1024. ViT features (especially when we are combining multiple layers of features together) contain a mix of redundant geometric information and high-level semantic signals. A simple linear projection forces a global transformation on all these features. In contrast, the use of a gated projection allows the network to choose to suppress noise or irrelevant/duplicate dimensions before they reach the narrow VAE bottleneck, making the VAE's task easier. In addition, the gated projection provides a linear path for gradients while giving more expressive power.

\subsection{The Power Spherical Distribution}
The von Mises-Fisher (vMF) distribution is the classical choice for spherical VAEs \cite{hyperspherical_vae}. However, its density depends on the modified Bessel function of the first kind $\mathcal{I}_v(\kappa)$, which lacks a closed-form inverse and complicates the reparametrization trick, often needing inefficient rejection sampling methods instead. To avoid this issue, we instead use the Power Spherical distribution \cite{power_spherical}. For a latent vector $\textbf{z} \in \mathbb{S}^{d-1}$ on the $(d-1)$-dimensional sphere, the probability density function is defined as:
\begin{equation}
q(\mathbf{z}; \mu, \kappa) = \frac{1}{C_{d}(\kappa)} (1 + \mu^\top \mathbf{z})^\kappa
\end{equation}
where $\mu \in \mathbb{S}^{d-1}$ is the location (mean) parameter, $\kappa > 0$ is the concentration parameter (inverse variance), and the normalization constant is given by:
\begin{equation}
    C_{d}(\kappa) = 2^{\alpha + \beta} \pi^\beta \frac{\Gamma(\alpha)}{\Gamma(\alpha + \beta)}
\end{equation}
with $\alpha = \kappa + \frac{d-1}{2}$ and $\beta = \frac{d-1}{2}$. This distribution is uniquely suited for deep learning because it admits a differentiable sampling procedure via the \textit{inverse transform method}.

\paragraph{Reparametrized Sampling}
In order to sample \mbox{$\textbf{z} \sim \text{PowerSpherical}(\mu, \kappa)$}, we first sample a scalar $y \sim Beta(\alpha, \beta)$. We then transform $y$ into a coordinate $t = 2y-1$ along the axis of $\mu$. A random vector $\textbf{v}$ is sampled uniformly from the $(d-2)$-dimensional sphere. The intermediate vector $y' = [t;\textbf{v}\sqrt{1-t^2}]$ represents a sample centered at the north pole $\textbf{e}_1 = [1, 0, \cdots, 0]^\top$. To align this sample with the predicted mean, we apply a Householder reflection:
\begin{equation}
    \mathbf{z} = \left( \mathbf{I} - 2\mathbf{u}\mathbf{u}^\top \right) \mathbf{y}', \quad \mathbf{u} = \frac{\mathbf{e}_1 - \mu}{|\mathbf{e}_1 - \mu|}
\end{equation}
This allows the gradient to flow through $\mu, \kappa$ without requiring numerical approximations of Bessel functions.

\subsection{Product of Spheres For High Dimensions}
A significant challenge arises when applying spherical distributions to high-dimensional latents ($d > 20$). The surface area of the $(d-1)$-dimensional sphere, $S_{d-1} = \frac{2\pi^{d/2}}{\Gamma(d/2)}$ vanishes exponentially as $d$ increases. This leads to numerical instability in the KL divergence and posterior collapse where the concentration $\kappa$ fails to converge. 

To represent a total latent dimension $D$, we instead define the latent manifold as a product of spheres:
\begin{equation}
    \mathcal{M} = \prod_{i=1}^N \mathbb{S}^{d'-1}, \quad \text{where } N \times d' = D
\end{equation}
Under this formulation, the encoder predicts $N$ independent pairs of $(\mu_i, \kappa_i)$. The total latent representation is the concatenation $\textbf{z} = [\textbf{z}_1, \cdots, \textbf{z}_N]$. Here, there is a tradeoff between number of spheres and the dimensionality of each sphere. While lower dimensional spheres are more stable and easier to train, products of spheres do not have the same topology as a single sphere (recall that the torus $T^2$ is topologically the same as the product $S^1 \times S^1$, but the torus has a hole that the individual factors do not have). Meanwhile, as we increase the dimensionality of each sphere, the numerical stability of the training decreases. As a good balance, we primarily use $d' \in \{ 4, 8\}$ and $N = D/d'$, which we found worked well empirically.

\paragraph{KL Divergence}
Given a uniform prior $\mathcal{U}(\textbf{S}^{d-1})$ on each sphere, the total KL divergence is the sum of the divergences for each sub-sphere. The KL divergence for a single sphere is the difference between the negative entropy of the learned distribution and the entropy of the uniform prior:
\begin{equation}
    D_{KL}(q | p) = -H(q) + \log\left(\frac{2\pi^{d'/2}}{\Gamma(d'/2)}\right)
\end{equation}
where the entropy $H(q)$ is computed using the log-normalizer and the digamma function $\psi$:
\begin{equation}
    H(q) = \log C_{d'}(\kappa) - \kappa \left( \log 2 + \psi(\alpha) - \psi(\alpha + \beta) \right)
\end{equation}

\subsection{Geometric Intuition and ViT Alignment}
We empirically demonstrate that the standard Gaussian prior $\mathcal{N}(0, I)$ is ill-suited for the task of compressing ViT features due to a fundamental topological mismatch. ViT tokens, constrained by Layer Normalization and utilized via a dot-product attention, reside approximately on a hyperspherical manifold $\mathbb{S}^{d-1}$ where semantic identity is encoded based on angular orientation instead of magnitude. The superiority of these hyperspherical VAEs can be broken down into three parts: (1) Topological Alignment, (2) Concentration of Measure, and (3) Semantic Consistency.

\paragraph{Topological Alignment}
The Gaussian prior assumes that the data occupies a solid Euclidean volume in $\mathbb{R}^n$. Meanwhile, LayerNorms constrain features to a specific mean and variance, effectively performing an approximate projection onto a high-dimensional hyperspherical shell. By trying to map data living on a sphere into a solid ball, this creates a topological problem. The model must waste latent capacity learning to ignore the radial dimension. By using a Power Spherical bottleneck, we align the latent prior's topology with the data's natural manifold.

\paragraph{Concentration of Measure}
In high dimensions, Gaussian distributions behave counter-intuitively. Specifically, as the dimension $d$ increases, the mass of a Gaussian $\mathcal{N}(0, I)$ concentrates almost entirely in a thin shell of thickness $O(1)$ within distance $\sqrt{d}$ from the origin. Thus, even though the probability density is the highest at the origin, the volume grows so quickly that there is effectively $0$ probability of ever sampling a point near the origin. 

While this sounds similar to the spherical VAEs, the Gaussian KL loss introduces a tension. Specifically, given some latent $\textbf{z}$, we can decompose it implicitly as a direction $\theta$ and a magnitude $r$. When we apply the standard KL divergence loss:
\begin{equation}
\mathcal{L}_\text{KL} = -\frac{1}{2}\sum_{j=1}^d (1 + \log (\sigma_j^2) - \mu_j^2 - \sigma_j^2)
\end{equation}
This loss tries to pull the mean $\mu$ towards the origin, and force the variance $\sigma^2$ towards $1$. This radial pressure towards the origin forces the model to choose between mapping features into the shell (and increasing the KL loss) or collapsing them towards the origin (which increases reconstruction loss). This can lead to posterior collapse.

Meanwhile, in the spherical VAE, we remove the radial dimension completely. This eliminates all of the radial noise, and allows the model to focus only on the important part, angular concentration. Additionally, on the surface of a high-dimensional sphere, two random vectors are almost certainly orthogonal. The spherical VAE can leverage this property to keep the latents separable. 

The product of spheres provides an additional compositional inductive bias. By partitioning the latent space into $N$ independent sub-manifolds, we allow the model to represent partial semantic overlaps - where tokens may share specific attributes in some sub-spheres while remaining distinct in others. 

\paragraph{Semantic Consistency}
The semantic meaning of a ViT token is defined in terms of its relationship to other tokens, which can be measured using cosine similarity. In Gaussian space, two vectors can have a very high cosine similarity while still being considered very different, if their magnitudes differ. Meanwhile, in a spherical VAE, the magnitude is fixed. Therefore, Euclidean distance and cosine similarity become equivalent.
\section{Experimental Setup}

\subsection{Datasets}
To ensure an appropriate mix of scene dynamics and camera motion, we train and evaluate the model on RealCam-Vid \cite{realcamvid}. The dataset is curated from three source datasets (RealEstate10K  \cite{realestate10k}, DL3DV-10K \cite{dl3dv10k}, and MiraData \cite{miradata}), that are refined through a multi-stage filtering pipeline. This collection yields a diverse set of scenes, including indoor and outdoor static scenes, plus real-world and animated/game environments which are dynamic. This allows us to verify that our VAE is able to handle all the types of scenes properly.

\subsection{Metrics}
We evaluate our approach using downstream task-specific metrics, depending on the backbone and task under consideration.

\paragraph{Depth Estimation Metrics}
For depth prediction tasks, we follow standard evaluation protocols and report Absolute Relative Error (AbsRel). We additionally report threshold accuracy $\delta_1$, which measures the fraction of pixels whose predicted depth is within a factor of 1.25 of the ground-truth depth. Additionally, for further exploration on the DINO models, we report Squared Relative Error (SqRel) and RMSE(log), which provide complementary views of depth estimation quality, with SqRel emphasizing large deviations and RMSE(log) capturing errors in log space. 

\paragraph{Camera Pose Metrics}
For camera pose estimation, we report Absolute Trajectory Error (ATE), which measures the deviation between predicted and ground-truth camera trajectories after alignment. This metric captures accumulated geometric drift and is sensitive to temporal consistency across frames. We additionally present AUC@30, defined as the area under the pose-accuracy curve up to a 30$^\circ$ rotation threshold. 

\paragraph{Point Cloud Metrics}
For point cloud reconstruction, we evaluate geometric fidelity using Chamfer Distance (CD) on confidence-filtered point clouds. Chamfer Distance measures the average bidirectional distance between predicted and ground-truth point sets.

\subsection{Training Details}
All our large VAEs are trained on either 4 x Nvidia A40s (46 GB) or 4 x Nvidia A100 (64 GB). For the VGGT-based models, we trained for 1 day. For the DINO models, we trained for 12 hours. We use the same hyperparameters across all models for consistency. We use $4$ input frames at a time, at a resolution of $224 \times 224$ (this primarily matters for the VGGT tests, as the DINO models will process the frames independently). Regardless of the input dimensionality of the features, we project the concatenated features of all dimensions into a $1024$-dimensional hidden space, and use a $128$-dimensional spherical bottleneck, as a product of $16$ spheres each having dimension $8$. The size of the VAEs range from 400M-500M parameters, depending on the input feature dimensionality.

Our VAE consists of $12$ sets of attention layers in the encoder and decoder, and $16$ attention heads per layer. We use bf16 mixed precision for all trainings. For optimization, we use the AdamW \cite{adamw} optimizer with a weight decay of 0.05, and a linear warmup to a learning rate of $1e-4$ and with cosine annealing decay. We also default to the initial $\kappa$ value of $30.0$ as a good starting point, but allow the model to learn the correct predictions for $\kappa$. 
\subsection{Loss Functions}
Our training losses can be split into two categories: the key losses to make the VAE work, and the task-specific losses. Here, we emphasize one key detail: for our training, we consider the original decoder head's outputs to be the ground truth. Since we are freezing all of the underlying models and decoder heads, comparing against a real ground truth leads to 2 potential sources of error: (1) reconstruction error caused by the VAE incorrectly reconstructing the features, and (2) inaccuracies caused by the decoder head itself. To remove this second error, we just use the original model outputs as the ground truth.

\paragraph{Main Loss Functions}
For our feature reconstruction losses, we use a combination of MSE and cosine similarity loss. Specifically, given the input features $\textbf{x}$ and the reconstructed features $\hat{\textbf{x}}$, our feature reconstruction loss is given $\mathcal{L}_\text{FR} = \text{MSE}(\textbf{x}, \hat{\textbf{x}}) + 0.25*\text{SIM}(\textbf{x}, \hat{\textbf{x}})$. 

While this works decently well by itself, we find that it is also important to preserve the relational structure between the different features. For this purpose, we define a loss on the Gram matrices of the feature layers. If $\textbf{X}$ is the $L_2$-normalized features of a given layer, we define the Gram matrix of these features as $\textbf{G} = \textbf{X}\textbf{X}^\top$. We then compute cosine distance between the Gram matrices of the original and reconstructed features. This forces the VAE to maintain inter-feature relation patterns, helping to preserve the overall structure of the scene even if the absolute feature values are compressed.

Furthermore, to counteract the tendency of variational bottlenecks to produce over-smoothed representations, we add two additional regularization losses. First, we use a variance-preserving loss, penalizing the $L_1$ distance between the variance of the input features and reconstructed features. Additionally, we apply a similar loss to the norms of the input and reconstructed features. 

Finally, as described above, we apply the KL divergence loss across the product of spheres. 

\paragraph{Task-Specific Losses}
For the task-specific losses, we use the original losses found in each work. For clarity, we briefly review the relevant losses here.

For VGGT, we apply the camera, depth, and point cloud losses found in the original model. The camera loss compares the predicted cameras $\hat{\textbf{g}}_i$ with the original cameras $\textbf{g}_i$ using a Huber loss: $\mathcal{L}_\text{Cam} = \sum_{i=1}^N ||\hat{\textbf{g}}_i - \textbf{g}_i||_{\epsilon}$. 

Meanwhile, the depth loss implements the aleatoric-uncertainty loss found in both VGGT \cite{vggt} and DUSt3R \cite{dust3r}. In this setup, the model produces both the depth map and an uncertainty map $\hat{\Sigma}_i^D$. The depth loss is computed as the difference between the predicted depth $\hat{D}_i$ and the original depth $D_i$, weighted by the uncertainty map. Additionally, VGGT also applies a gradient term, which we use too. The final depth loss is given by:
\begin{equation}
\begin{split}
\mathcal{L}_\text{Depth} &= \sum_{i=1}^N ||\hat{\Sigma}_i^D \odot (\hat{D}_i - D_i)|| \\ &+ ||\hat{\Sigma}_i^D \odot (\nabla \hat{D}_i - \nabla D_i)|| - \alpha \log \hat{\Sigma}_i^D
\end{split}
\end{equation}
The pointmap is defined similarly, using a pointmap uncertainty $\hat{\Sigma}_i^P$.

\section{Results}

\subsection{Reconstruction Quality for VGGT}

\begin{figure*}
    \centering
    \includegraphics[width=0.9\linewidth]{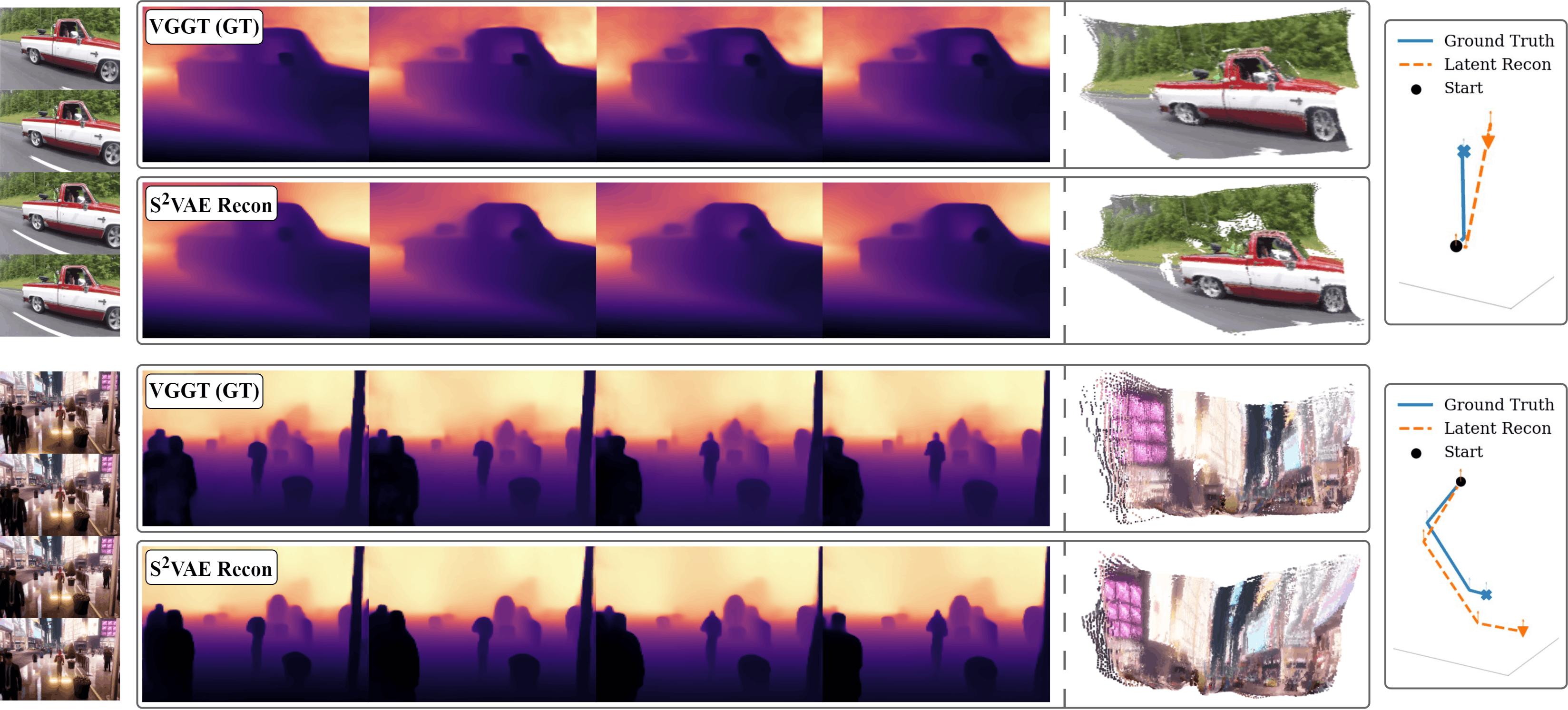}
    \caption{\textbf{Qualitative reconstruction of geometry from compressed latent representations.}}
    \label{fig:reconstruction_quality}
\end{figure*}

Here, we qualitatively explore the performance of our VAE in reconstructing the task-specific outputs of the VGGT model. Figure \ref{fig:reconstruction_quality} shows a sample set of reconstructions, involving depth map, camera trajectory, and point cloud. We can see that our VAE is able to reconstruct all the different tasks to a high degree of accuracy, without the presence of many artifacts.

\begin{table}[h]
    \centering
    \caption{\textbf{Transformer backbones evaluated in this work}, along with their parameter counts and associated downstream tasks. VGGT is evaluated on depth, camera pose, and point cloud reconstruction, DINOv2 variants are evaluated on depth estimation, DUSt3R is evaluated on depth and point clouds, and CLIP is evaluated based on final generated samples via Stable Diffusion. }
    \begin{tabular}{lcc}
    \hline
        \textbf{Model} & \textbf{Num. Params.} & \textbf{Tasks} \\
        \hline
        DINOv2-B & 86M & Depth \\
        DINOv2-L & 300M & Depth \\
         VGGT & 1.2B & Depth + PC + Camera \\ 
         DUSt3R & 600M & Depth + PC \\
         CLIP Text & 63M & Image Generation \\
         \hline
    \end{tabular}
    
    \label{tab:compared_models}
\end{table}

\subsection{Exploring Other ViT Architectures}
While VGGT is the primary focus of this work, its geometry-aware training may make certain tasks inherently easier. To assess the generality of our approach, we therefore evaluate it across a range of ViT architectures and model scales. In particular, we consider the DINOv2~\cite{dinov2} family, including DINOv2-Base, DINOv2-Large, and DINOv2-Giant.
Table~\ref{tab:compared_models} summarizes the ViT backbones considered in our evaluation, along with their model sizes and the downstream tasks used for assessment.

Quantitative depth estimation results across these models are reported in Table~\ref{tab:depth_results}.

\begin{table}[!t]
    \centering
    \caption{\textbf{Quantitative depth estimation results for different ViT backbones using the proposed VAE framework.} Lower AbsRel and higher $\delta_1$ scores indicate better performance.
}
    \begin{tabular}{lcccc}
    \hline
        \textbf{Model} & \textbf{AbsRel} & $\bm{\delta_1}$ & \textbf{SqRel} & $\bm{\mathrm{RMSE}}_{\log}$ \\
        \hline
        DINOv2-B & 0.035 & 0.99 & 2.8e-4 & 0.05 \\
        DINOv2-L & 0.037 & 1 & 3.4e-4 & 0.04 \\
         \hline
    \end{tabular}
    
    \label{tab:depth_results}
\end{table}

\subsection{Exploring the Spherical Latent Space}
In order to demonstrate that our model has actually learned a meaningful latent space, we perform spherical interpolation between two sets of latents. Recall that spherical interpolation between latents $\textbf{x}$ and $\textbf{y}$ is given by the following formula:
\begin{equation}
    \text{Slerp}(\textbf{x}, \textbf{y}, t) = \frac{\sin((1-t)\Omega)}{\sin(\Omega)} \textbf{x} + \frac{\sin(t\Omega)}{\sin(\Omega)}\textbf{y}
\end{equation}
where $\Omega = \arccos(\textbf{x} \cdot \textbf{y})$ is the angle between the two spherical vectors.

\begin{figure}[h]
    \centering
    \includegraphics[width=\linewidth]{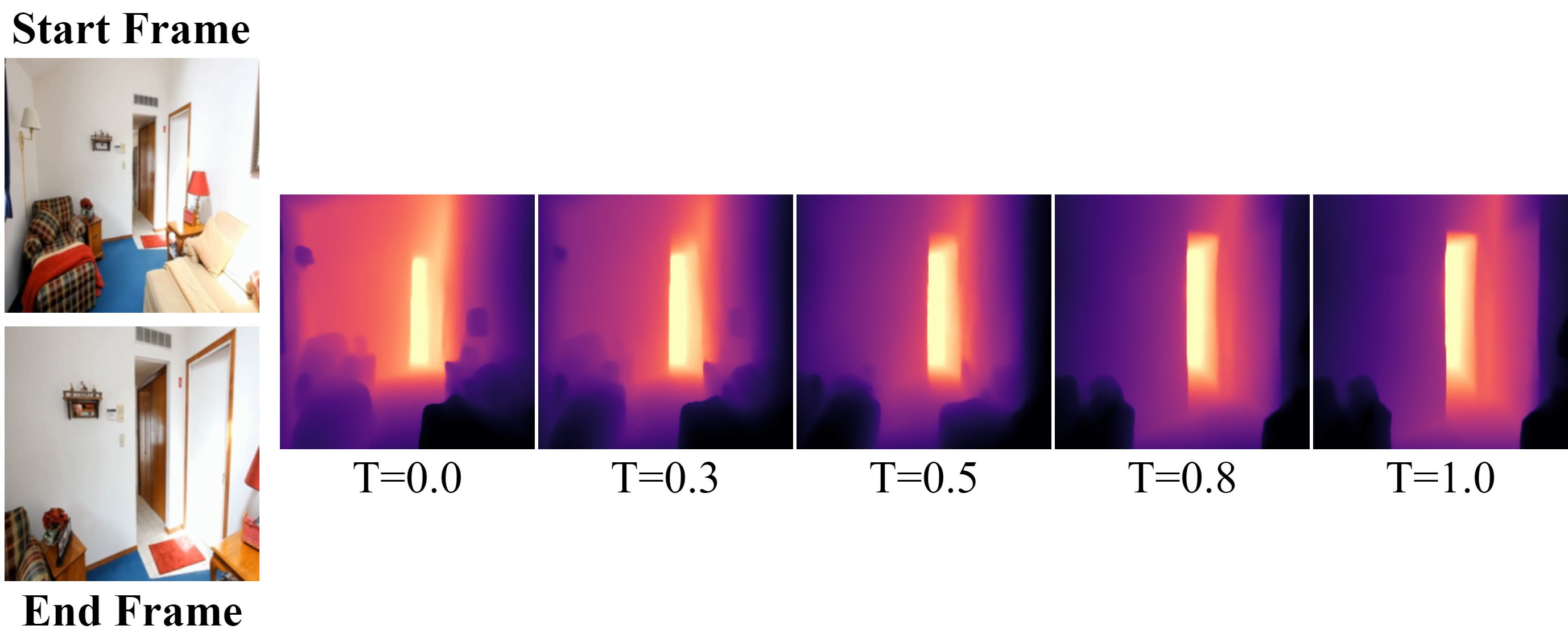}
    \caption{\textbf{Spherical Interpolation.} Demonstrating the effectiveness of our latent space, by performing spherical interpolation between two sets of latents from different scenes. We see a reasonably smooth transition between the two depth maps.}
    \label{fig:spherical_interpolation}
\end{figure}

Figure \ref{fig:spherical_interpolation} shows an example in terms of the predicted depth maps of the first frame, as we interpolate between two completely different scenes. We can see that we get quite smooth transitions between the two depth maps, and no sudden jumps or sharp changes. This suggests that our spherical latent space is also meaningful and informative, even at low KL weights ($10^{-4})$. 

\subsection{Ablations}
\begin{table*}[h!]
    \centering
\caption{\textbf{Ablation results for bottleneck distribution choice on the VGGT architecture.}}
    \begin{tabular}{lccccc}
        \hline
        \textbf{Ablation} & \textbf{AbsRel} ($\downarrow$) & $\bm{\delta_1}$ ($\uparrow$) &
        \textbf{CD} ($\downarrow$) &
        \textbf{ATE} ($\downarrow$) & \textbf{AUC@30} ($\uparrow$) \\
        \hline
         P.S. Many Spheres (Ours) & \textbf{0.04} & \textbf{0.98} & \textbf{0.31} & \textbf{0.001} & \textbf{99.5} \\
         Gaussian & 0.07 & 0.95 & 0.34 & 0.002 & 99.3 \\
         P.S. One Sphere & 1.21 & 0.25 & 0.32 & 0.002 & 99.4 \\
         \hline  
    \end{tabular}
    \label{tab:bottleneck_choice}
\end{table*}
We ablate several design decisions of our architecture. The most important such decision is the choice to use a spherical distribution instead of a Gaussian distribution or some other distribution.

Specifically, we explore the use of several different distributions: (1) Gaussian distribution, (2) Power spherical with one sphere, and (3) Power spherical with many spheres. Table \ref{tab:bottleneck_choice} shows the results on the VGGT architecture across all metrics. As can be seen in this table, the product of many spheres outperforms both the gaussian as well as single, high-dimensional sphere. Notably, the performance of the single sphere is very poor due to the numerical instability in high dimensions.

\subsection{Preliminary Demonstration on DiTs}
While the primary focus of this work is the proposed VAE, we include a preliminary experiment to illustrate its applicability to downstream generative models. Specifically, we train a text-conditioned Diffusion Transformer (DiT) directly in the compressed latent space to generate VGGT features, which can subsequently be rendered into depth maps, camera poses, and point clouds. We emphasize that this experiment serves solely as a proof of concept, and that neither the architecture nor the training procedure was optimized for this setting.

We employ a standard DiT~\cite{dit} augmented with cross-attention layers for text conditioning, using T5-XXL~\cite{t5} as the text encoder. Since the model operates on vector-valued latents rather than spatial grids, we apply the DiT without patchification. Training is performed using rectified flow~\cite{rectified_flow}, with the generated latents projected back onto the hyperspherical manifold.

\begin{figure}[h]
    \centering
    \includegraphics[width=\linewidth]{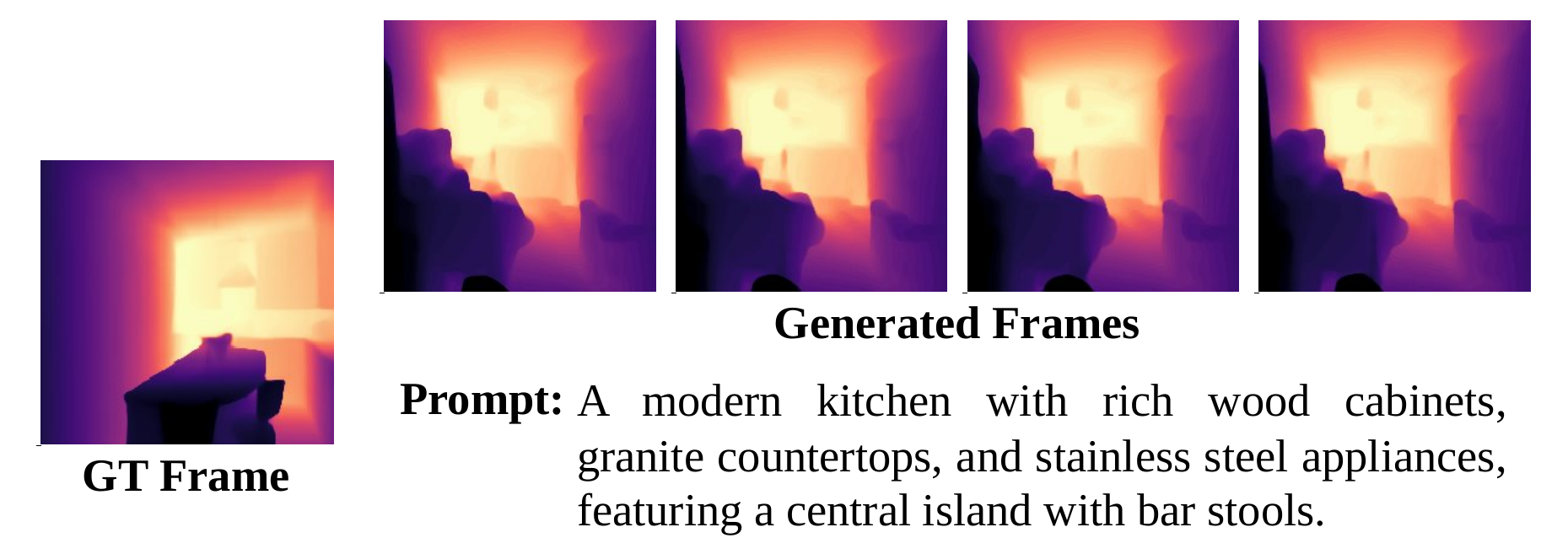}
    \caption{\textbf{Qualitative results after preliminary training of a DiT model.} Despite not tailoring the DiT architecture or training procedure to spherical latents, the model produces reasonable text-to-latent generations, which can be rendered into depth maps using the VGGT decoder heads.}
    \label{fig:diffusion_results}
\end{figure}

Figure~\ref{fig:diffusion_results} shows qualitative results after one day of training on four A40 GPUs. The resulting generations exhibit temporal consistency and coherent geometric structure across frames, indicating that the proposed latent space can support diffusion-based generation. Future work will explore DiT architectures and training strategies specifically tailored to hyperspherical latents.

\section{Conclusion}
We investigated the compression of high-dimensional ViT features, arguing that the standard Gaussian latent assumption fails to capture their intrinsic geometry. Driven by normalization and attention mechanisms, ViT features reside on approximately hyperspherical manifolds where semantic information is encoded directionally. We demonstrated that Euclidean Gaussian bottlenecks introduce systematic failures, such as posterior collapse and semantic drift, that become particularly acute in geometry-sensitive tasks.

To resolve this, we introduced a geometry-aligned VAE utilizing a product of hyperspherical latent distributions. By explicitly modeling directional structure, our approach preserves angular semantics and ensures stability without forcing the model to implicitly learn spherical constraints. Experimental results across various backbones show that this alignment yields superior reconstruction fidelity and downstream performance, especially under high compression. Our findings suggest that latent geometry is a critical design factor for scaling transformer-based representations, and future work will explore its further integration with diffusion models and other non-Euclidean structures.
\section*{Impact Statement}
This paper presents a methodological contribution aimed at improving the compression and reuse of transformer-based representations by aligning latent variable models with the intrinsic geometry of ViT features. The primary goal of this work is to advance the understanding and design of representation learning techniques in machine learning.

The proposed approach focuses on model architecture and latent space design and does not introduce new data sources, application domains, or deployment mechanisms. As such, we do not foresee immediate ethical concerns or societal risks arising uniquely from this work beyond those commonly associated with the development of general-purpose machine learning models.

More broadly, improved representation compression may contribute to more efficient and reusable models, potentially reducing computational and energy costs in large-scale learning systems. We believe these outcomes are generally beneficial and aligned with ongoing efforts toward more sustainable and robust machine learning research.

\bibliography{example_paper}
\bibliographystyle{icml2026}

\newpage
\appendix
\onecolumn
\section{Additional VGGT Experimental Details}

\subsection{Active Dimensions}
Gaussian bottlenecks have a tendency to utilize only a percentage of their latent dimensions. To demonstrate that our spherical VAE utilizes these dimensions better, we pick 1000 random samples from our dataset, obtain the latents for each sample, and check the variance across the sequence. We find that on average, for a Gaussian bottleneck trained on $128$ dimensions, only $91$ of the dimensions are active (we define active to be a dimension having a variance of at least $0.1$). Meanwhile, all $128$ of the spherical dimensions exhibit at least $0.1$ variance, indicating that the spherical dimensions are being utilized much more effectively compared to the Gaussian ones. Additionally, the variance was much more uniform for the spherical distribution, with all dimensions having between $0.1$ and $0.15$ variance. Meanwhile, the Gaussian distribution had dimensions ranging from much below $0.1$, all the way up to $3.5$ variance. This indicates overutilization of certain dimensions while ignoring others.

\subsection{Required Dimensionality to Compress VGGT}
Another important question is the number of dimensions required to actually compress the VGGT features and obtain meaningful reconstructions. To determine this, we run experiments using our spherical VAE on a wide range of latent dimensions. For each experiment, we use a hidden dimension of $128$, and $8$ layers each with $8$ heads. In total, all the models had between $9.1$M and $9.3$M parameters.

We can see in Table \ref{tab:latent_dim} that even as we decrease the latent dimension to extremely small values ($8$), we are able to achieve quite high results in all metrics, and in fact the $8$ latent dimension model performs the best in depth estimation. On the reverse side, the bigger models perform slightly better on the point cloud and camera pose metrics. Given the high reconstruction performance, this can potentially indicate that the very large hidden dimensions of the VGGT model are not properly being utilized, or contain a lot of redundant information. Future work will seek to explore this in far more detail.

\begin{table}[h!]
    \centering
    \caption{\textbf{VGGT Latent Dimension Experiments.} A comparison of our VAE's performance at reconstructing the VGGT features, as we fix a low model size and vary the latent dimension. In all experiments, we use 8 dimensions per sphere (so 64-dimensional latents would use 8 spheres).}
    \begin{tabular}{lcccccc}
    \hline
         \textbf{Latent Dim} & \textbf{Num Params} & \textbf{AbsRel} $(\downarrow)$ & $\bm{\delta_1} (\uparrow)$ & \textbf{CD} $(\downarrow)$ & \textbf{ATE} $(\downarrow)$ & \textbf{AUC@30} $(\uparrow)$ \\
         \hline
         8 & 9.164M & \textbf{0.07} & \textbf{0.95} & 0.35 & 0.005 & 99.3 \\
         16 & 9.169M & 0.09 & 0.92 & 0.33 & 0.004 & 99.3\\
         32 & 9.177M & 0.09 & 0.94 & \textbf{0.32} & 0.004 & 99.3\\
         64 & 9.194M & 0.10 & 0.86 & 0.33 & \textbf{0.003} & 99.3 \\
         128 & 9.227M & 0.10 & 0.93 & \textbf{0.32} & \textbf{0.003} & \textbf{99.4} \\

         \hline
    \end{tabular}
    \label{tab:latent_dim}
\end{table}

\section{Additional Theory}
Here, we elaborate more on the challenges that Gaussian bottlenecks face compared to spherical bottlenecks in compressing these features. Specifically, I focus on a few key details.

\subsection{Exploding Gradients for Gaussian Bottlenecks}
Consider the case of antipodal points $x_1, x_2$ on the sphere, so that $||x_1 - x_2||_2 = 2$. Their latent representation in terms of the gaussian bottleneck will approximately lie on the ball $B_{\delta}$ for some $\delta$, depending on the KL loss. But then, any decoder function $g$ will have approximately Lipschitz constant given by $L \geq \frac{1-\epsilon}{\delta}$, where $\epsilon$ is the desired reconstruction error bound. Thus, as the KL loss goes to $0$, we will see $\delta \to 0$, implying that the Lipschitz constant grows arbitrarily large. This can lead to large instabilities during training.

\subsection{Approximate Spherical Property of Layer Normalization}
While the ideal LayerNorm would always project the features onto the sphere, this is not true due to the inclusion of the $\epsilon$ term used for numerical stability. However, we can approximately characterize what probability a feature has of being close to the sphere, depending on $\epsilon$. Specifically, if the minimum variance of the dimensions is bounded below by some $\sigma_{min}$ with probability $1 - \delta$, then with this same probability, the output of the LayerNorm will lie on a shell of thickness approximately $\frac{\sqrt{d} \epsilon}{2 \sigma_{min}}$. Thus, as long as we have variance across all the dimensions, we can attain a very thin shell on which most features will be contained.

\subsection{Numerical Verification of the Spherical Property}
We also seek to verify numerically that this spherical property actually does hold in practice. To do this, we run experiments across both VGGT and DINOv2, using $1000$ test samples, and calculate the mean and variance of the norms across each layer. Note that pre-LN transformer architectures tend to see the magnitude of the features increasing with depth (due to the unbounded residual).

We find that across both models and all layers, the coefficient of variation (CV) of the norms remains under $0.15$, indicating that most of the norms are clustered quite strongly around the mean of that layer. This justifies why the spherical approximation is valid for these sets of features. 

Specifically, for the VGGT model, all layers except the last layer have CV of $0.07$ or $0.08$, while the last layer does jump up to $0.13$. Howver, these are still low enough that the hyperspherical assumption remains valid. For the DINOv2-Base model, we see a much more concentrated set of means, decreasing from $0.08$ at the first layer to $0.06$ in the last layer. Even the randomly initialized version of DINOv2 follows a similar pattern, with CV $0.09$ at the first layer, and decaying down to also $0.06$ in the last layer. This indicates that this is not a training-specific property, but instead holds more generally due to the LayerNorm usage.

\section{Tests on Different Models}
To further validate the capabilities of our architecture, we test on other types of transformer-based models, not just DINO-based (we consider VGGT to be DINO-based since it uses as input the DINOv2 features). 

\subsection{CLIP Text Encoder}
CLIP \cite{clip} is a widely used text-image model, trained on unifying image and text representations in a single latent space. To do this, the CLIP model has two separate encoders (one for images, one for text). However, the text encoder is the most widely used, found in a wide range of generative models \cite{styleclip, vidstyleode, stablediffusion}, as well as zero-shot image classification, multi-modal retrieval \cite{clip_multimodal_retrieval} and open-vocabulary segmentation \cite{segment_anything}.

We seek to understand how well our VAE architecture is able to reconstruct the CLIP textual features and apply them for a downstream task. Specifically, we want to use these reconstructed textual features for text-to-image generation, via the Stable Diffusion \cite{stablediffusion} model. We focus on the v1 version of stable diffusion as it uses only CLIP for the text encoding, compared to Stable Diffusion 3 \cite{stablediffusion3} which also uses a T5 \cite{t5} encoder as well (this T5 will prevent us from truly measuring the image quality, as it will not be clear what part of the information comes from our reconstructed CLIP features).

For the architecture, we use the same setup as the VGGT and DINOv2 case, except with a hidden size of $512$ instead of $1024$. The model size is $\approx 105M$ parameters. We train only on the main loss functions (using no task-specific losses). We train for $50$ epochs on the same dataset (RealCam-Vid) using the provided text captions.

\begin{figure}
    \centering
    \includegraphics[width=\linewidth]{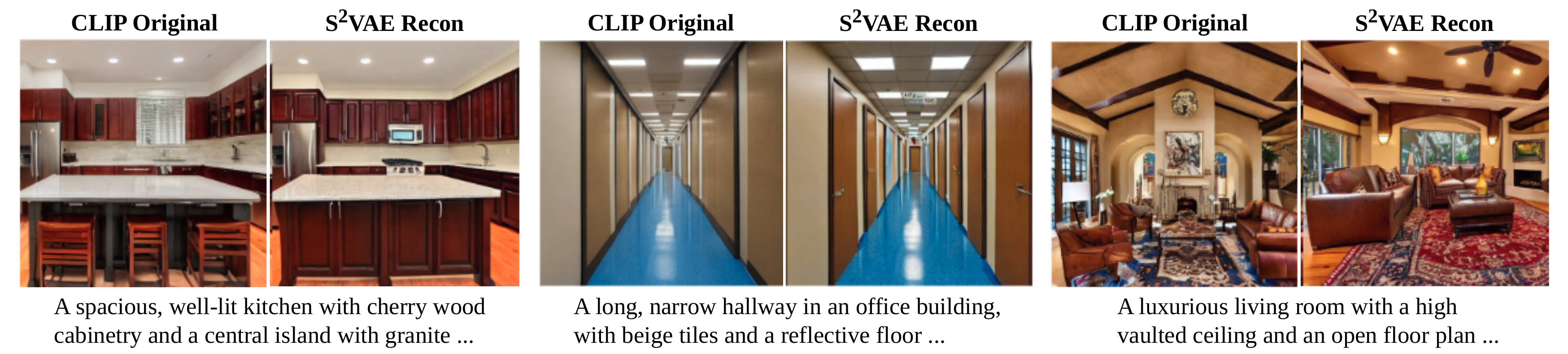}
    \includegraphics[width=\linewidth]{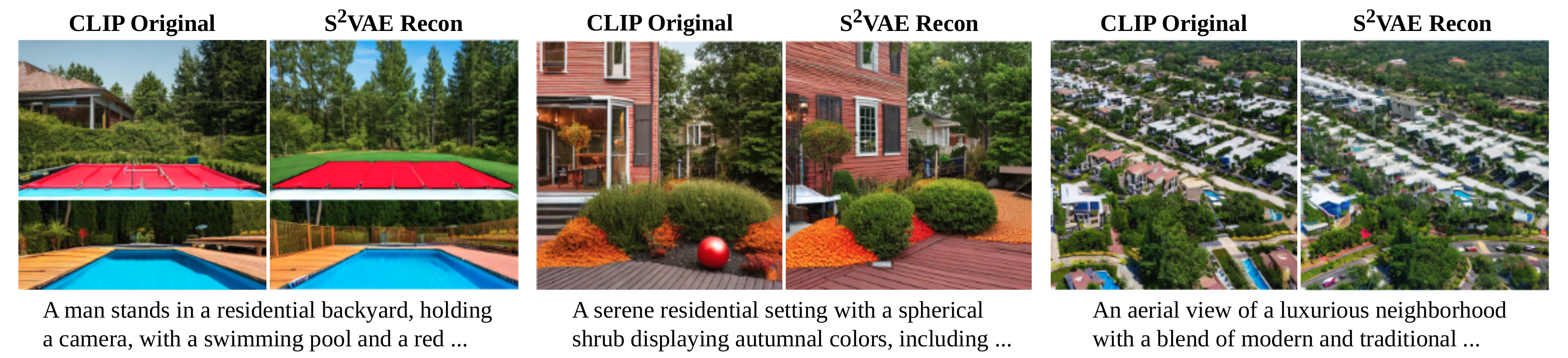}
    \caption{Images generated by the Stable Diffusion model using the original CLIP text features vs the features reconstructed using our VAE, using captions found in the test set of the RealCam-Vid dataset. We can see that all the key semantic details of the original images are always found in the reconstructed versions too, showing that our VAE is able to properly reconstruct all the relevant semantic information in the text features, even without any explicit task-specific losses.}
    \label{fig:clip_sd_results_in_domain}
\end{figure}

Figure \ref{fig:clip_sd_results_in_domain} shows reconstruction results comparing the original image that Stable Diffusion generates vs the generated images using the reconstructed CLIP features. We see that the reconstructed features are able to maintain all the key semantic details in the image, across a diverse set of scenes. Additionally, within this dataset, we get an average of $\approx 0.94$ cosine similarity between the true features and the reconstructed features, showing that the model is able to preserve almost all of the semantic information stored in the text features.

\begin{figure}
    \centering
    \includegraphics[width=\linewidth]{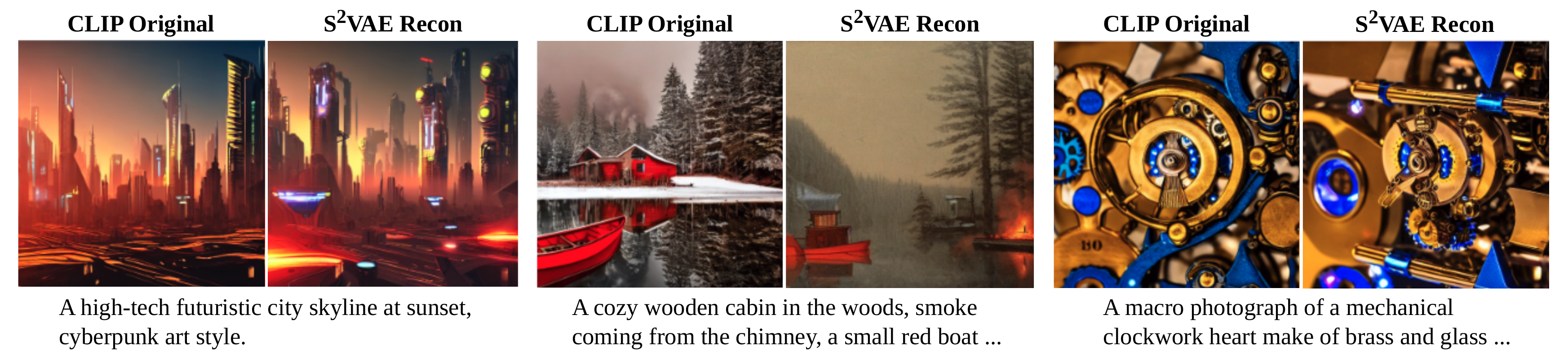}
    \caption{Several samples on randomly generated, out-of-distribution prompts. We can see that our model generalizes quite well, preserving the core semantics despite being trained only on RealCam-Vid samples.}
    \label{fig:clip_sd_results_ood}
\end{figure}

We also test on out-of-domain captions, specifically random captions generated by an LLM. We see that there is still a high level of semantic similarity between the original and reconstructed images, although some of the very specific details are a bit different. However, we still get an average cosine similarity of $\approx 0.82$, which is sufficient for preserving the most important details of the image.

\begin{figure}
    \centering
\begin{subfigure}[b]{0.415\textwidth}
\includegraphics[width=\linewidth]{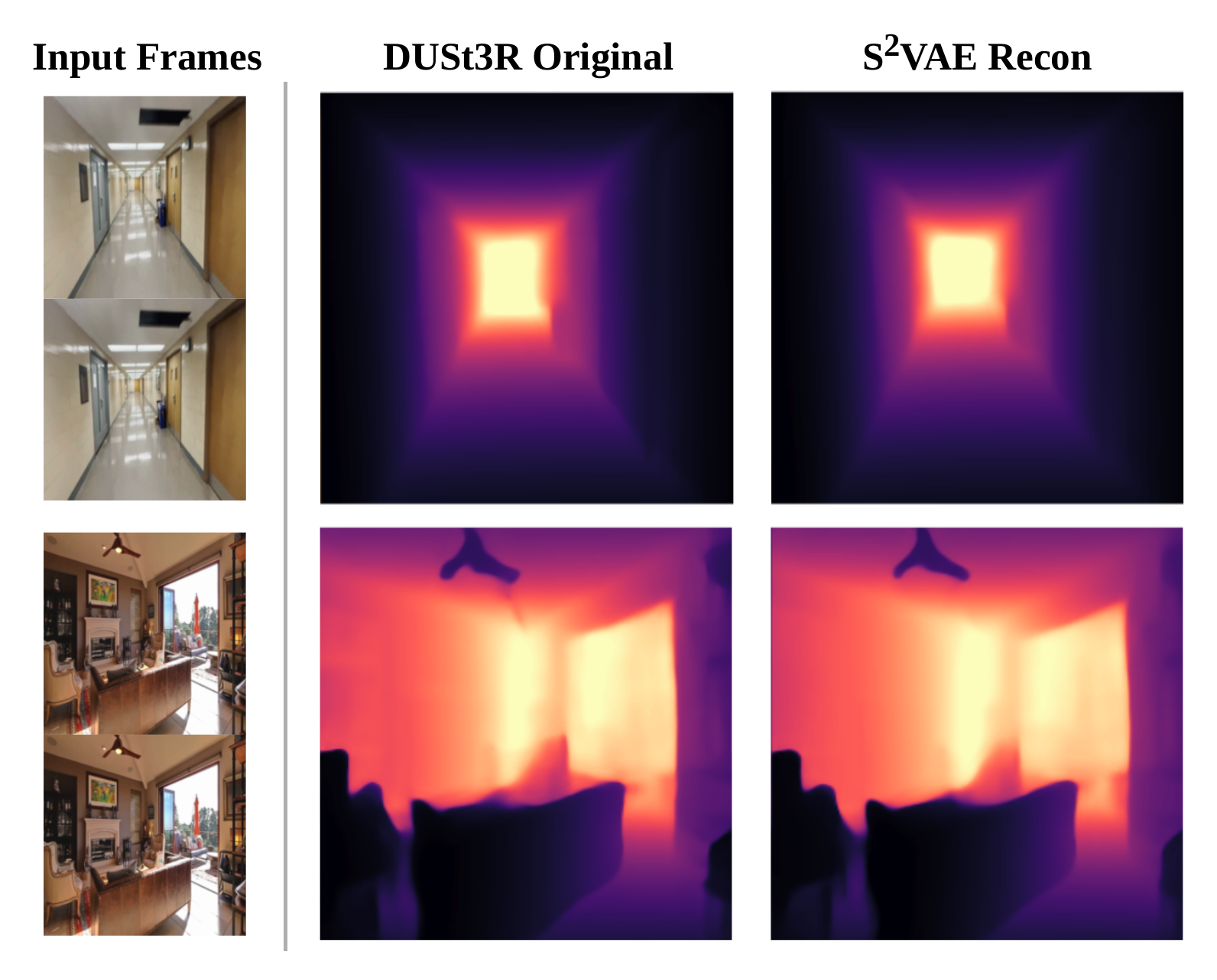}
    \caption{Frame pairs from the RealCam-Vid test set, with corresponding original depth maps from DUSt3R and their reconstructions from our S$^2$VAE. Our method achieves near-perfect reconstruction.}
    \label{fig:dust3r_results}
    \end{subfigure}
    \hspace{10pt}
    \begin{subfigure}[b]{0.545\textwidth}
    \includegraphics[width=\linewidth]{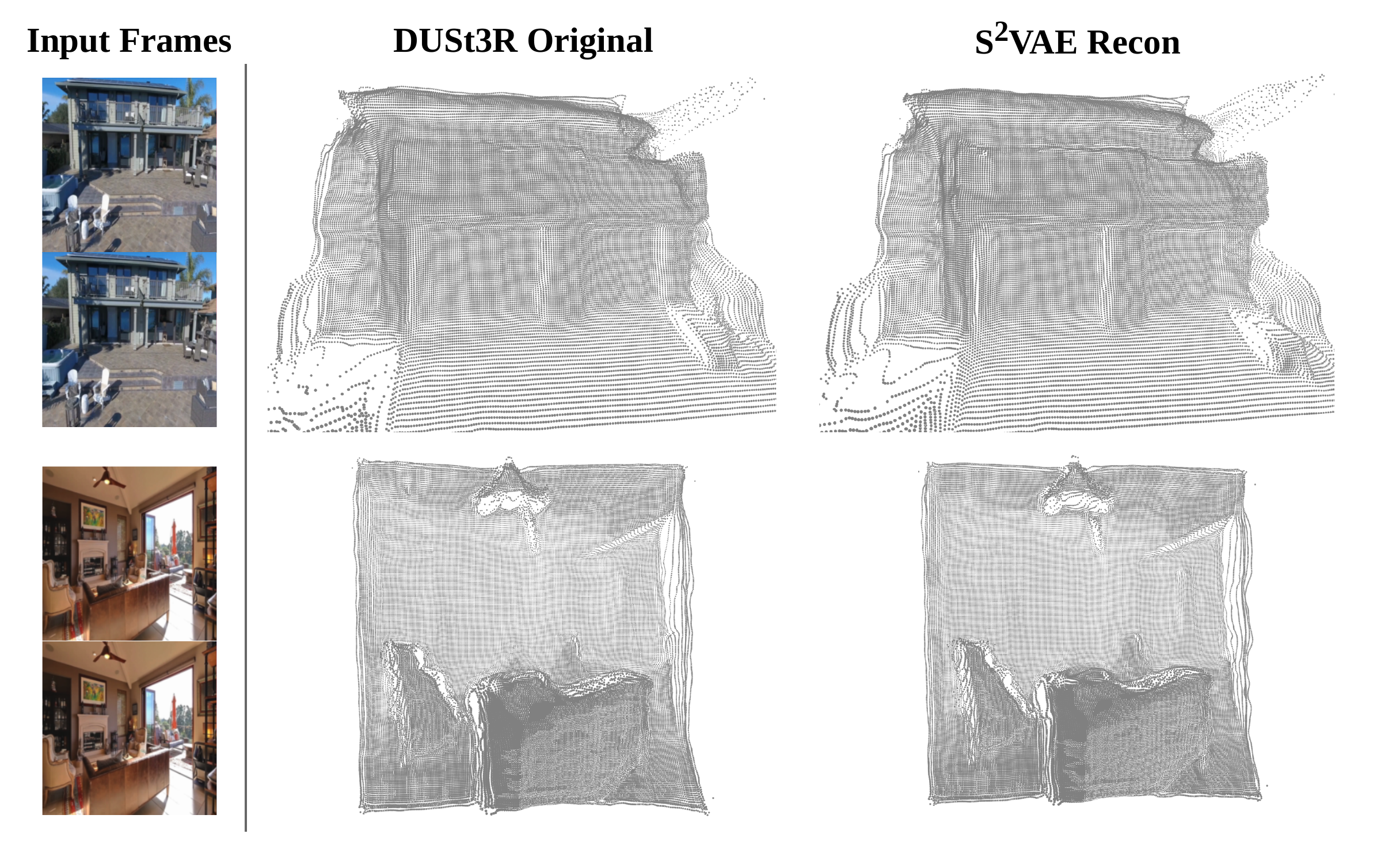}
    \caption{Frame pairs from the RealCam-Vid test set, with corresponding point cloud outputs from DUSt3R and their reconstructions from our S$^2$VAE. The reconstructed point clouds preserve both global structure and fine-grained details.}
    \label{fig:dust3r_results}
    \end{subfigure}
\end{figure}

\subsection{DUST3R Model (CroCo Backbone)}
DUSt3R \cite{dust3r} is a popular model for stereo 3d reconstruction, which processes pairs of images and attempts to reconstruct global pointmaps of the scenes, as well as the confidence levels and camera information (for the second image relative to the first image). To do this, DUSt3R relies on the CroCo \cite{croco} backbone, a self-supervised ViT trained on the masked image modeling objective to understand 3d geometry. Thus, this is again a very different training objective than DINO or VGGT. 

To test our reconstruction quality on the DUSt3R model, we focus on reconstructing the features of the CroCo backbone. Specifically, we take two images of a scene, and pass them through the CroCo model to obtain the two sets of features. We then treat each set of features similarly to how we treated each of the VGGT layer features, and concatenate the two sets of features together along the feature dimension (this allows the VAE model to rely on information from both views at the same time). 

We use the same setup as the CLIP version, using the RealCam-Vid dataset, and train for $50$ epochs. The model size is again around $\approx 108M$ parameters.
\vspace{-3mm}
\section{Sequence-Level Compression}
While the previous experiments included in the paper keep the number of tokens the same as the original method while reducing the feature dimensionality, it is also possible to compress along the sequence length too. Specifically, we adapt a method similar to \cite{flatdino}. In this method, in order to compress from $N$ tokens per frame to $M < N$ tokens per frame, we randomly initialize and learn $M$ register tokens to be used for the compression. These $M$ tokens are appended to the front of our input sequence (and duplicated based on the number of frames), and then the model performs the same self-attention operations as before. Then, the bottleneck layer uses only these $M$ register tokens per frame and discards all the original input tokens.

Meanwhile, for the decoder side, we require learning another $N$ register tokens, and perform the same strategy, appending to the front of the sequence of latents (with duplication if necessary), performing the self-attention layers, and then using only those tokens as the reconstructed features. By doing this, we are able to achieve both strong feature-level and sequence-level compression. 

For simplicity, we use the same number of layers as before, with a fixed latent dimension of $128$ and a fixed hidden dimension of $1024$ for all these experiments. Thus, the only variable between experiments is the number of tokens we use per frame. We can see in Table \ref{tab:register_tokens} the performance based on the number of register tokens used. Even when we use only $8$ register tokens per frame, we are still able to achieve quite strong reconstruction metrics across all tasks. We believe that this is due to our model being able to exploit the significant amount of redundancy across the tokens to achieve heavy compression without much loss. Additionally, since we still do require $N$ register tokens to be appended for the decoder, this can help during the reconstruction.

\begin{table}
    \centering
    \begin{tabular}{lccccc}
    \hline
    \textbf{Num Register Tokens} & \textbf{AbsRel} $(\downarrow)$ & $\bm{\delta_1} (\uparrow)$ & \textbf{CD} $(\downarrow)$ & \textbf{ATE} $(\downarrow)$ & \textbf{AUC@30} $(\uparrow)$  \\
    \hline
     8 & 0.039 & 0.976 & 0.175 & 0.0008 & 0.997\\
     16 & 0.043 & 0.97 & 0.173 & 0.0008 & 0.996\\
     32 & 0.036 & 0.978 & 0.174 & 0.0009 & 0.996\\
     64 & 0.035 & 0.976 & 0.183 & 0.0007 & 0.997\\
     128 & 0.029 & 0.986 & 0.176 & 0.0007 & 0.996\\
    \hline
    \end{tabular}
    \caption{Number of register tokens used per frame for sequence-level compression and their corresponding metrics. Interestingly, while we see steady improvement in the depth metrics as we increase the number of register tokens, both the point cloud and camera metrics stay mostly the same across all compression levels, implying that the point cloud information is more heavily prioritized by the model even at heavy compression levels, compared to depth.}
    \label{tab:register_tokens}
\end{table}
\vspace{-3mm}
\section{Failure Cases}
\begin{figure}
    \centering
    \includegraphics[width=0.70\linewidth]{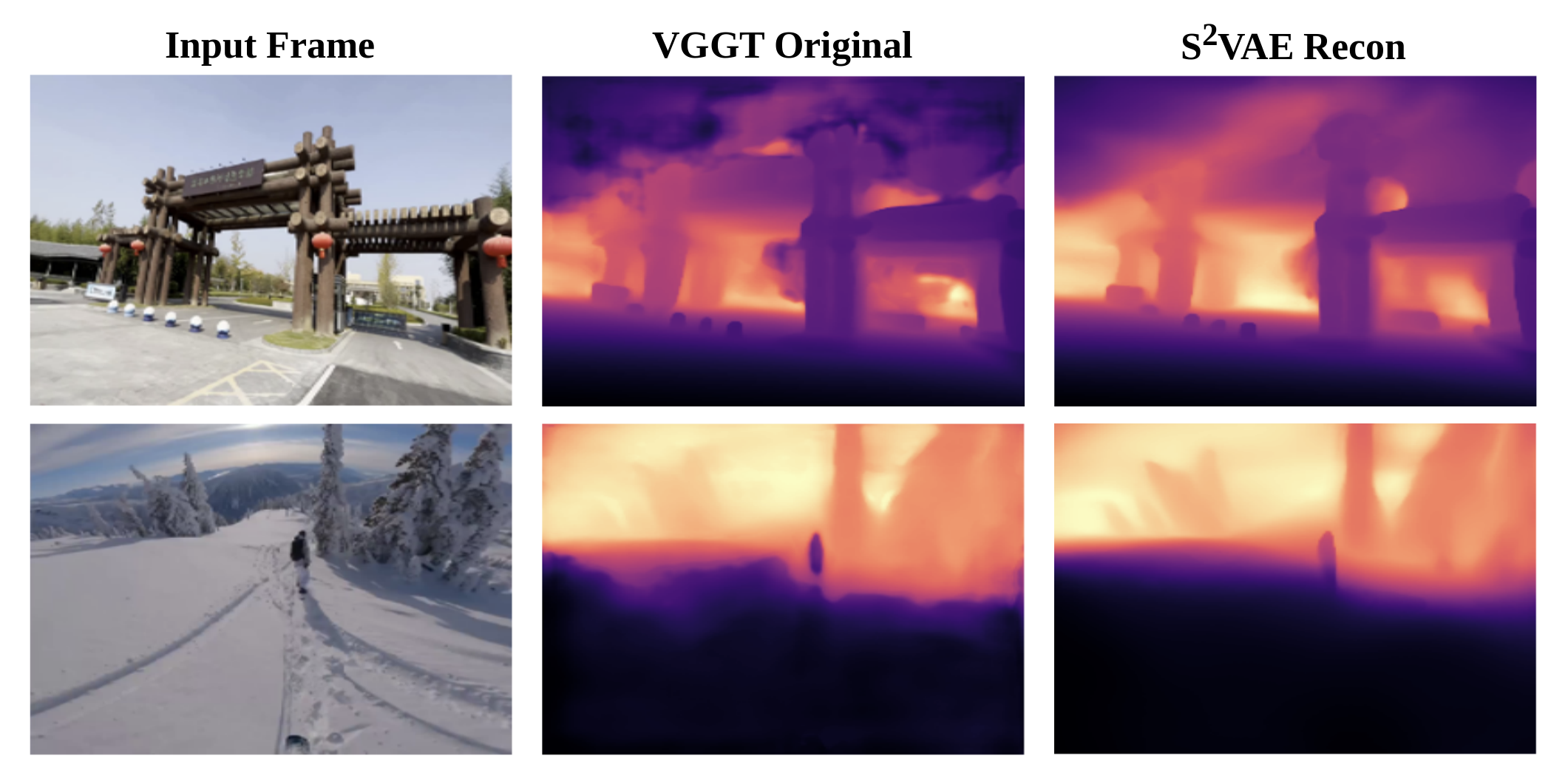}
    \caption{Several examples where our model fails to properly reconstruct more complex textures in the depth map, leading to a heavily smoothed result. However, even in this case, we still manage to strongly preserve the overall structure and content.}
    \label{fig:failure_cases}
\end{figure}

We also perform an analysis of certain failure cases of our method. Specifically, as expected due to the heavy rate of compression we use, the model has a tendency to oversmooth certain textures. Figure \ref{fig:failure_cases} shows several examples where we can see significantly more detail in the depth maps produced by VGGT normally, compared to our reconstruction. While this is less impactful for depth maps compared to regular images, it is still worth noting.

\section{Lightweight Probing}
\begin{figure}
    \centering
    \includegraphics[width=\linewidth]{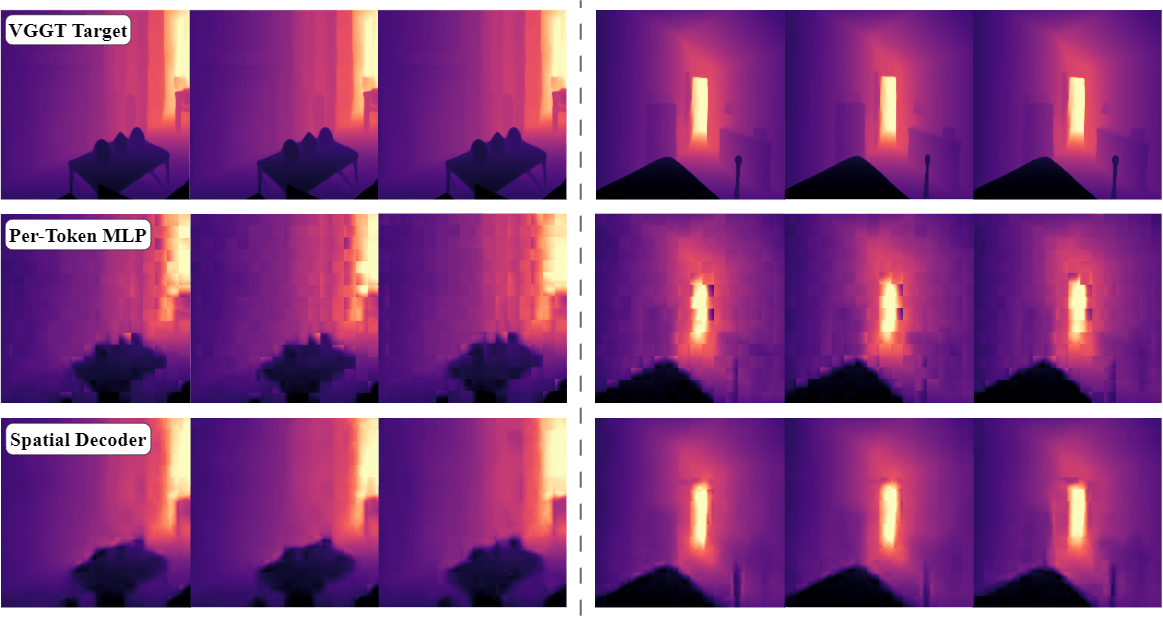}
    \caption{Several samples using our lightweight probes directly on the VAE latent space. Both the per-token, two-layer MLP and the two-layer spatial (CNN) decoder are able to reconstruct the depth maps while containing all of the key objects and relative distances, while both being under 500k parameters. The MLP version has grid artifacts due to not being able to share information among tokens, but the CNN version smoothes this out and just lacks some of the finer detail. This shows that our latents are learned sufficiently well that the task information is easily accessible and usable.}
    \label{fig:probe}
\end{figure}

In order to demonstrate that the learned latent space is directly meaningful, we train several lightweight probes to convert the VAE latents directly into a depth map. Specifically, we test both a two-layer per-token MLP (which directly converts a token into a 16x16 patch of the output depth map), as well as a two-layer spatial (CNN)-based decoder. Both methods contain under $500$k parameters, and do not have any interaction with the VGGT decoders. Evaluated on a $1000$ video test set, we find that the MLP has an AbsRel of $0.065$, and this is reduced by the CNN decoder to $0.049$. Figure \ref{fig:probe} shows a few samples of this. The MLP has grid artifacts due to the lack of interaction between tokens before the output, and the CNN smoothes these artifacts out but lacks some of the higher-frequency detail. However, in both cases, the core structure and objects remain in the scene, showing that even a weak probe is able to extract all the meaningful depth information from the latents.

\section{Sub-sphere Specialization}
One important question relating to our product-of-spheres formulation is how the important information is distributed among the different spheres, and whether the spheres implicitly specialize without any specific losses. In order to verify this, we computed Spearman correlation and mutual information between the each of the $16$ sub-spheres in our base model (having $16$ spheres of dimension $8$ each, for a total latent dimension of $128$) over $200$ validation samples. 

We find that certain spheres encode point-level geometry (with $|\rho| = 0.65-0.74$ among these spheres). Meanwhile, other spheres carry over $2\times$ the average mutual information for camera information. Depth seems to be distributed more broadly than the other properties, with most spheres having some depth-specific information, but a few containing also over $2\times$ the average mutual information. Critically, no sphere was completely uninformative, with $|\rho| > 0.19$ for all spheres and tasks. 

This demonstrates that the model learns to utilize certain spheres for containing more information about a specific task, without explicit supervision. In addition to the stability improvements that the product-of-spheres formulation provides, these results show that the formulation also helps the model distribute information more evenly and effectively.

\section{Cohesive Generation Capabilities}
\begin{figure}
    \centering
    \includegraphics[width=0.5\linewidth]{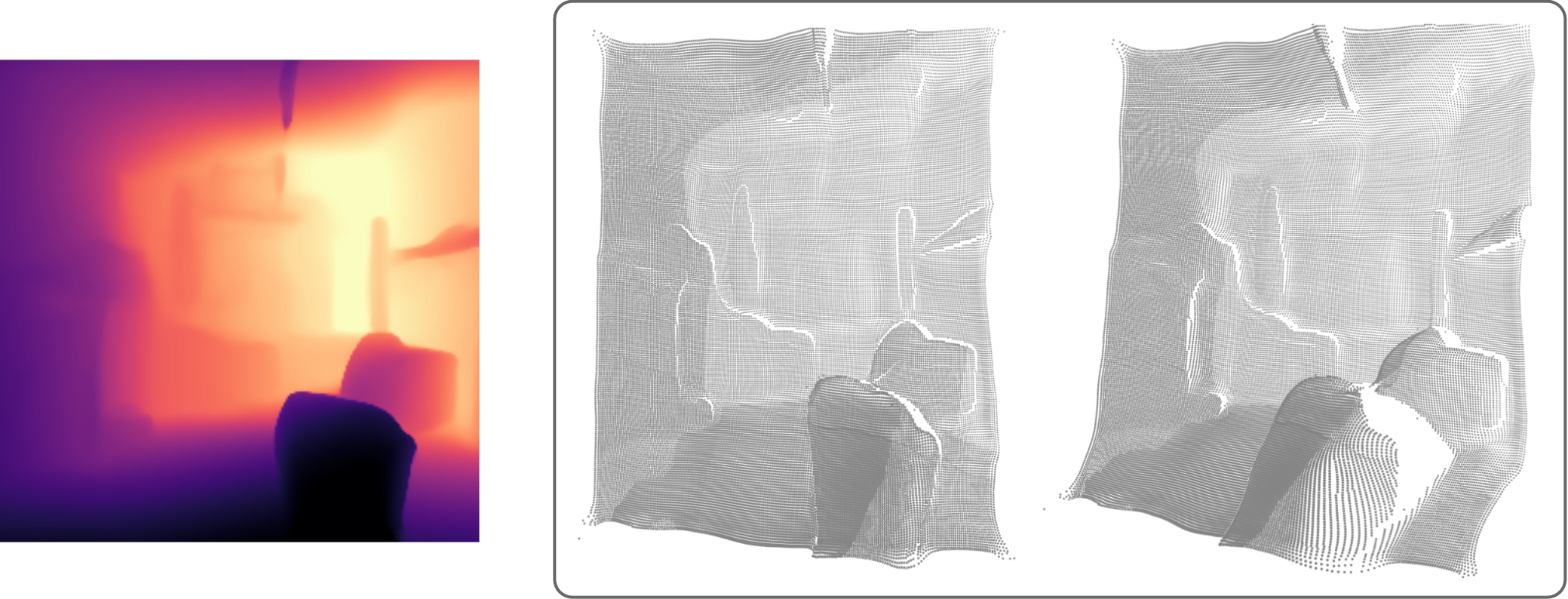}
    \caption{By using the joint feature space of the VGGT model for generation, we are able to ensure that the generated depth maps and point clouds agree with each other, even for completely generated scenes. Here, we see the generated depth map for one sample, and the corresponding generated point cloud (viewed from two different angles). Note that the objects in the point cloud scene are arranged exactly like the depth map, including their depth from the camera in this view.}
    \label{fig:gen_depth_pc}
\end{figure}

By enabling generative models to work on compressed transformer features, we enable joint cohesive generation across multiple modalities at once. Specifically, VGGT features are used jointly for depth, point cloud, and camera trajectories. Therefore, by generating these features directly, we can enable generations which agree across all modalities. Figure \ref{fig:gen_depth_pc} shows an example generated sample, where both the depth map and point cloud completely agree, including all the objects and their depth relative to the camera. This opens up new possibilities of training generative models with multiple types of outputs, by relying on the strength of pretrained transformer models.

\end{document}